\pdfoutput=1

\documentclass[11pt]{article}

\usepackage[preprint]{acl}

\usepackage{times}
\usepackage{latexsym}
\usepackage{amsfonts}

\usepackage[T1]{fontenc}

\usepackage[utf8]{inputenc}

\usepackage{microtype}

\usepackage{inconsolata}

\usepackage{graphicx}

\title{The BIAS Detection Framework: Bias Detection in Word Embeddings and Language Models for European Languages}

\author{Alexandre Puttick, Leander Rankwiler, Catherine Ikae, Mascha Kurpicz-Briki \\
  Applied Machine Intelligence \\
  Bern University of Applied Sciences \\
  Biel/Bienne, Switzerland \\
  \texttt{alexandre.puttick@bfh.ch, catherine.ikae@bfh.ch, mascha.kurpicz@bfh.ch} }

\begin{document}
\maketitle
\begin{abstract}
The project \emph{BIAS: Mitigating Diversity Biases of AI in the Labor Market} is a four-year project funded by the European commission and supported by the Swiss State Secretariat for Education, Research and Innovation (SERI). As part of the project, novel bias detection methods to identify societal bias in language models and word embeddings in European languages are developed, with particular attention to linguistic and geographic particularities. This technical report describes the overall architecture and components of the \emph{BIAS Detection Framework}. 
The code described in this technical report is available\footnote{https://github.com/BFH-AMI/BIAS} and will be updated and expanded continuously with upcoming results from the BIAS project. The details about the datasets for the different languages are described in corresponding papers at scientific venues.
\end{abstract}

\section{Introduction}
The matter of \emph{bias} has become one of the major limitations of the latest applications of machine learning. With the increasing use of machine-learning-based language models, the issue of bias in the field of natural language processing (NLP) is of great concern.  \citet{hovy2021five} highlight five dimensions along the machine learning pipeline in which bias can be introduced: the data, the annotation process, the input representations, the models, and the research design. The dimension of input representations refers chiefly to so-called word embeddings, which encode human language as mathematical vectors. This includes the study of the output of the hidden layers of transformer-based large language models.

A large body of work targets the detection and mitigation of bias in word embeddings and language models; for an overview see e.g., \citet{meade2022empirical, sun2019mitigating, delobelle2022measuring}. An early detection and mitigation method was developed by \citet{bolukbasi2016man}, attempting to delete gender information from word embeddings by identifying and contracting the vector subspace corresponding to gender. Another well-known and often used method for bias detection in word embeddings, the Word Embedding Association Test (WEAT), was proposed by \citet{caliskan2017semantics}. Since then, researchers have applied, reviewed, adapted and built-upon these methods in various settings. The effectiveness of these methods in mitigating bias has also been called into question \citep{gonen2019lipstick}. While early methods were designed for static word embeddings, methods for contextualized word embeddings have since been developed, e.g., \citet{may2019measuring,kurita2019measuring, nangia2020crows,ahn2021mitigating, guo2021detecting}. Nevertheless, the majority of research focuses on the English language. The bias encoded in mathematical models often originates from the societal stereotypes of a given culture. Recent work in the field has indicates that such bias can depend on the language and the cultural context, e.g., \citet{kurpicz2020cultural} \citet{kurpicz2021world}. Additional research efforts, such as those described in this technical report, are therefore a necessity.

What is considered as an unwanted \emph{bias} in the data or outcome of machine learning models involves not only technical but also societal aspects. The discussion of bias in NLP research has been criticized for being vague, inconsistent, and lacking in normative reasoning \cite{blodgett-etal-2020-language}. In this interdisciplinary project, we tackle this challenge by collaborating closely with experts from the domains of social sciences, humanities and law.

In this work we present the \emph{BIAS Detection Framework}, which extends state-of-the-art methods for bias detection in word embeddings and language models to several European languages, with regional and cultural aspects discovered and developed within our interdisciplinary research consortium. The work presented in this technical report was executed in the context of the Horizon Europe project \emph{BIAS}, which is described in more detail in the next section. We release the results concerning the bias detection and mitigation methods for word embeddings and language models in European languages in a public Github repository\footnote{https://github.com/BFH-AMI/BIAS}. This report gives an overview of the architecture of the code base for bias detection. Specific results for the different languages will be published at scientific venues. Please refer to the README file for implementation and further details.

\section{The BIAS Project}
The Horizon Europe project \emph{BIAS Mitigating Diversity Biases of AI in the Labor Market}\footnote{https://www.biasproject.eu/} brings together an interdisciplinary consortium of nine partner institutions to develop an understanding of the use of AI in the employment sector and how to detect and mitigate unfairness and bias in AI-driven recruitment tools. In particular, as part of the technical work package, several tasks are dedicated to the investigation of new bias detection and mitigation methods for static word embeddings and language models. Leveraging the knowledge from an interdisciplinary consortium (including, for example, literature about societal stereotypes in the different regions, interviews, surveys), and co-creation activities, these tasks study how human bias is reflected in AI technology. In the context of these activities, interdisciplinary co-creation workshops were conducted in the different partner countries, including experts from HR, NGOs and AI. The project also aims to raise awareness and provide learning materials for different stakeholder groups, complementing bias mitigation at a technical level--which poses an immense challenge--by addressing bias at the organizational and human level.

\section{The BIAS Detection Framework: System Architecture}

\begin{figure*}[ht]
   \centering
    \includegraphics[width=.8\textwidth]{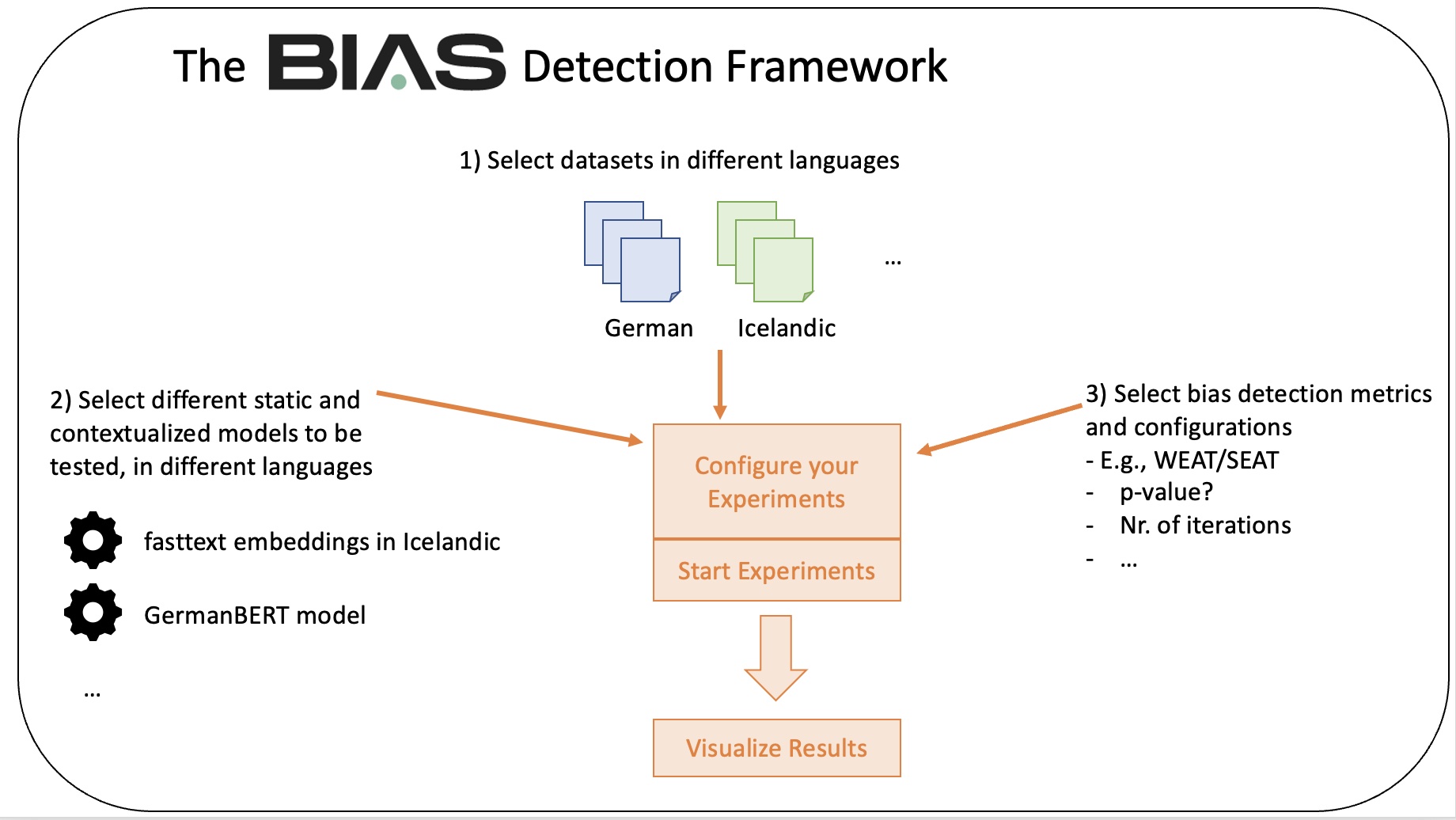}
    \caption{Overview of the BIAS Detection framework.}
    \label{fig:framework}
\end{figure*}

Figure \ref{fig:framework} gives an overview of the functionalities of the framework. The framework supports different types of models, including fasttext\footnote{https://fasttext.cc/} or BERT-based models via the HuggingFace transformer library\footnote{https://huggingface.co/docs/transformers/index}. Note that the models are not included in the repository and need to be downloaded as needed. Refer to the README file for detailed instructions.

The different bias detection methods from the state-of-the-art mostly rely on word lists or sentence templates. Those datasets are available per language, and can be selected for the configuration of the experiments. Note that more datasets will be made available continuously and updated in the README file of the repository.

Batches of experiments can then be defined, by selecting models in combination with datasets and the bias detection metrics that will be explained more in detail in the next section. 

After execution of the experiments, the results are made available in a logfile, pre-formatted as Latex tables and in a graphical visualization. 

\section{The BIAS Detection Framework: Bias Detection Methods}
In this section, we describe each of the bias detection methods that we have implemented. An in-depth comparison of these methods for the English language can be found in \citet{delobelle2022measuring}.

\subsection{WEAT}
The Word-Embedding Association Test (WEAT) \citep{caliskan2017semantics} is inspired by the Implicit Association Test (IAT) \citep{greenwald1998measuring}, which is used is Psychology research to measure implicit bias in human subjects. Each WEAT test requires two categories of wordlists: \textit{attributes} and \textit{targets.} The attributes consist of wordlists $A$ and $B$ representing opposing concepts relating to an aspect of social bias. For example, $A$ = \{\textit{executive, management},\ldots\} and $B=$~\{\textit{home, parents,}\ldots\} are attribute lists representing the concepts of \emph{career} and \emph{family} respectively. The targets are also wordlists $X$ and $Y.$ In the context of social biases, those can represent different values of a particular sensitive trait, e.g., $X =$ \{\textit{male, man,}\ldots\} and $Y=$ \{\textit{female, woman,}\ldots\} in the case of gender. Using these wordlists, the WEAT test offers a quantitative measure of the degree of bias present in the word embeddings being studied. For example, given the \textit{career vs. family} and \textit{male vs. female terms}, the WEAT test provides a single number measuring the extent to which men are more closely associated to career, whereas women are more closely associated to family. In the original WEAT paper, eight tests related to social bias are defined, which are detailed in Table \ref{tab:weat_tests}.
\begin{table*}[ht]
    \centering
    \resizebox{\textwidth}{!}{
    \begin{tabular}{c|c|c|c}
         \textbf{Test} & \textbf{Bias Type} & \textbf{Targets} & \textbf{Attributes} \\
          WEAT 3-5$^*$ & ethnic/racial & European American vs. African American first names & pleasant vs. unpleasant \\
          WEAT 6 & gender & male vs. female first names & career vs. family \\
          WEAT 7 & gender & math vs. arts & male vs. female terms \\
          WEAT 8 & gender & science vs. arts & male vs. female terms \\
          WEAT 9 & health & physical vs. mental illness & temporary vs. permanent\\
          WEAT 10 & age & young vs. old people's names & pleasant vs. unpleasant\\
    \end{tabular}}
    \caption{A list of the WEAT tests that directly pertain to social biases. $^*$WEAT 4 uses a subset of names from WEAT 3, and WEAT 5 uses a subset of attributes from WEAT 4.}
    \label{tab:weat_tests}
\end{table*}

We will describe next how the WEAT-metric is computed in detail. Let $w$ be a word with corresponding word-embedding $\vec{w}$. The expression
$$
s(w,A,B) = \mathop{\mathrm{mean}}_{a\in A}\cos(\vec{w},\vec{a}) - \mathop{\mathrm{mean}}_{b\in B}\cos(\vec{w},\vec{b})
$$
measures to what extent $w$ is more closely associated to $A$ or $B$. The sign of $s(w,A,B)$ indicates the direction of the bias, while the magnitude indicates the level of bias. For example, if $w=$ \textit{man} and $A$ and $B$ correspond to \textit{career} and \textit{family} respectively, and the embedding space indeed encodes stereotypical bias, we would expect $s(w,A,B)$ to be a large positive number. The relative association between the target words $X,Y$ and the attribute words $A,B$ is then given by 
$$
s(X,Y,A,B) = \sum_{x\in X}\frac{s(x, A,B)}{|X|} - \sum_{y\in Y} \frac{s(y,A,B)}{|Y|}.
$$

The overall WEAT bias metric, called the \emph{effect size}, is computed by normalizing $s(X,Y,A,B)$:
\begin{equation}\label{eq: WEAT}
es(X,Y,A,B) = \frac{s(X,Y,A,B)}{\mathop{\mathrm{stddev}}_{w\in X\cup Y} s(w,A,B)}.
\end{equation}
Typically, $X,Y$, and $A,B$ are selected such that positive effect sizes indicate stereotypical bias, while negative values indicate anti-stereotypical bias. This is exemplified in the given scenarios comparing \emph{male vs. female} terms and \emph{career vs. family} attributes. The effect size may indicate the presence and magnitude of bias, however, it cannot be ruled out that the measured effect size is an artifact of the way word vectors are distributed in the embedding space. \citet{caliskan2017semantics} propose a significance test, the \textit{one-sided permutation test}, in order to ensure that random partitions of the target words $X\cup Y$ do not yield large spurious effect sizes. Let $\{X_i,Y_i\}_i$ denote the set of partitions of $X\cup Y$ into two sets of equal size. The $p$-value for the permutation test is given by
\begin{equation}\label{eq: permutation_test}
p := \mathrm{Pr}_i[s(X_i,Y_i,A,B)>s(X,Y,A,B)],
\end{equation}
i.e., the fraction of partitions for which $s(X_i,Y_i,A,B)>s(X,Y,A,B).$ A common threshold for statistical significance is $p < 0.05$, meaning that the null hypothesis (that there is no significant bias present) can be rejected at a $5\%$ level of significance.

\subsection{SEAT}
With the goal of detecting bias in contextual word embeddings, \citet{may2019measuring} introduced the Sentence Embedding Association Test (SEAT) as a simple method for extending WEAT. Target and attribute words are inserted into semantically bleached templates such as \textit{This is WORD} or \textit{WORD is here.} The word embeddings from WEAT are replaced with sentence embeddings, other than that SEAT scores are computed in the exact same manner as WEAT scores.
\subsubsection*{Embedding methods}
SEAT is intended to work with both static and contextual word embeddings, but the manner in which the sentence embeddings are obtained depends on the model being used. For example, for fasttext static embeddings, sentence representations are simply the average of the word vectors over the words in the sentence. For BERT and RoBERTa models, we have implemented several methods for obtaining the contextual word embedding associated to a given sentence:
\begin{itemize}
\item \texttt{[CLS]}: In the original SEAT implementation, \citet{may2019measuring} use the final hidden state of the \texttt{[CLS]} token as a sentence embedding.\footnote{RoBERTa models use the $\langle s\rangle$ token in place of \texttt{[CLS]}.}
\item \texttt{target\_first}: \citet{tan2019assessing} replace the \texttt{[CLS]} token with the embedding corresponding to the first subtoken of the WEAT word inserted into the SEAT sentence template.
\item \texttt{target\_pooled}: \citet{delobelle2022measuring} include the additional option of averaging the embedding vectors obtained from all sub-tokens of the target word.
\end{itemize}
For WEAT tests on BERT models, the model input is of the form '\texttt{[CLS]} X \texttt{[SEP]}'; thus the same variety of embedding methods can be used. This option is also implemented in the BIAS Detection Framework. Refer to the README for detailed instructions.

\subsection{Log Probability Bias Score (LPBS)}
The Log Probability Bias Score \citep{kurita2019measuring} is a WEAT-based bias metric specifically designed for masked language models (MLMs) such as BERT. The main insight of this work is to replace the use of cosine-similarity as a measure of the level of association between a target (e.g., \textit{man}) and an attribute (e.g., \textit{programmer}). Instead, the \textit{probability} the model estimates for the target and attribute to appear together in the same sentence is used. This method is only applicable to models trained using a masked language modeling objective, i.e., to predict masked tokens, denoted as \texttt{[MASK]}, given the token's context. For example, given an input of the form $x=$ \texttt{[MASK]} \textit{is a programmer}, the model will output a probability estimate $p(\mbox{\texttt{[MASK]}}=w\vert x)$, the probability that the masked token is given by the word $w$, for every word $w$ in the model's vocabulary.

To compute the association between the target \textit{male gender} and the attribute \textit{programmer}, first the probability that the sentence `\texttt{[MASK]} \textit{is a programmer}' will be completed with the word \textit{he} is computed. 
\begin{equation}
p_{tgt} = p\big(\mbox{\texttt{[MASK]} = \textit{he}}\big\vert \mbox{\texttt{[MASK]}\textit{ is a programmer}}\big)
\end{equation}
Independent of the context, the model may be statistically more or less likely to predict the word \textit{he} than the word \textit{she}, for instance if the corpus the model was trained on contains many more references to male subjects. To account for this difference and isolate the contribution of the word \textit{programmer} to the model's predictions, the probability

\begin{equation}
p_{prior} = p\big(\mbox{\texttt{[MASK]}$_1$ = he}\big\vert \mbox{\texttt{[MASK]}$_1$ \textit{is a} \texttt{[MASK]}$_2$}\big)
\end{equation}

is also computed and used to normalize $p_{tgt}.$ 

In general, the association between an arbitrary target $x$ and attribute $a$ is defined as
\begin{equation}\label{eq: log_prob_score}
asc(x,a) = \log\frac{p_{tgt}(x\vert a)}{p_{prior}(x)},
\end{equation}
where $p_{tgt}$ and $p_{prior}$ are computed exactly as in the above case with $x =$ \textit{he} and $a =$ \textit{programmer}. \citet{kurita2019measuring} refer to $asc(x,a)$ as the \textit{increased log probability score}. 
A positive association signifies that the likelihood of the target increases when the attribute is present, whereas a negative association indicates that the likelihood of the target decreases when combined with the attribute.

The increased log probability score $asc(x,a)$ is analogous to the cosine similarity $\cos(\vec{x},\vec{a}),$ and is used to compute an \textit{effect size} completely analogous\footnote{In the denominator in Eq. \ref{eq: log_WEAT}, the standard deviation is computed over attribute words, as opposed to Eq. \ref{eq: WEAT}. The main reason for this seems to be that the authors omitted many target words from their tests, most likely for issues in creating grammatically correct sentences. The resulting distribution does not capture the variance in $s_{log}$ values effectively.} to WEAT. Namely, given an attribute word $a$ and target word lists $X$ and $Y$, one computes:
$$
s_{\log}(a ,X,Y) = \mathop{\mathrm{mean}}_{x\in X}asc(x,a) - \mathop{\mathrm{mean}}_{y\in Y}asc(y,a),
$$
One defines $s_{\log}(X,Y,A,B)$ analogously and computes the effect size
\begin{equation}\label{eq: log_WEAT}
es_{\log}(X,Y,A,B) = \frac{s_{\log}(X,Y,A,B)}{\mathop{\mathrm{stddev}}_{w\in A\cup B} s_{\log}(w,X,Y)}.
\end{equation} 
For simplicity, we refer to this bias metric as \textit{LPBS}. The authors compute LPBS using the template \textit{TARGET is ATTRIBUTE} and target and attribute lists from WEAT tests 1, 3 and 6-8. To test statistical significance, the authors carry out a two-sided permutation test (t-test) over partitions of the combined attributes $A\cup B$, as opposed to WEAT, which uses partitions of $X\cup Y$. They show that LPBS indicates statistically significant bias in BERT that was not detected by conducting ordinary WEAT on contextual word embeddings extracted from the final hidden layer.

The original LPBS English-implementation was already limited by the difficulty of creating grammatically correct sentences from templates, a challenge which only increases with more grammatically complex and gendered languages. Within the BIAS Detection Framework, it was therefore decided as an initial step to use templates of the form \textit{TARGET ATTRIBUTE}. Although this does not result in complete sentences, the resulting LPBS effect sizes can still be useful; they can corroborate other bias measurements or potentially detect biases overlooked by other methods. 

\subsection{CrowS-Pairs}
The CrowS-Pairs bias metric is also designed specifically for BERT models, and aims to respond to criticism of the use of templates such as those in \citet{kurita2019measuring} for the limited scope and contrived nature of the resulting sentences. \citet{nangia2020crows} address this by compiling the \textit{Crowd-sourced Stereotype Pairs (CrowS-Pairs)} dataset, which consists of 1508 sentence pairs dealing with nine types of social bias: race, gender, sexual orientation, religion, age, nationality, disability, physical appearance and socioeconomic status/occupation. Each pair consists of a stereotypical sentence $s_s$, such as \textit{Fat people can never really be attractive}, and an antistereotypical counterpart $s_a$, e.g., \textit{Thin people can never really be attractive.} The sentences $s_s$ and $s_a$ are as semantically similar as possible, only differing in the terms representing the demographic groups being compared. As opposed to template-based methods, it is asserted that the crowd-sourced nature of the dataset results in greater diversity in both sentence structure and the stereotypes expressed. 

Similar to \citet{kurita2019measuring}, the authors use sentence pairs to define a probability-score-based bias metric. Given a sentence pair $(s_s,s_a)$, let $U$ be the set of shared words in $s_s$ and $s_a$, e.g., $U$ = \{\textit{people}, \textit{can}, \textit{never}, \textit{really}, \textit{be}, \textit{attractive}\}. Rather than using the increased log probability score (Eq. \ref{eq: log_prob_score}) to measure the likelihood of the sentence $s_s$, the authors adapt the \textit{psuedo-log-likelihood (PLL)} score \citep{salazar2019masked}
\begin{equation}\label{eq: pll}
    pll(s_s) := \sum_{u\in U}\log(p(\mbox{\texttt{[MASK]}} = u\vert s_s\setminus u),
\end{equation} where $s_s\setminus u$ denotes the sentence $s_s$ with a \texttt{[MASK]} token in place of the word $u$, e.g., \textit{Fat }\texttt{[MASK]}\textit{ can never really be attractive}. Using the above example concerning physical appearance, $pll(s_s)$ can be interpreted as the likelihood the model attributes to the remaining part of the sentence given the presence of the word \textit{fat} in the beginning.

The difference 
$$b_{s_s,s_a}^{p\log} := pll(s_s)-pll(s_a)$$
is then a bias measure quantifying the degree of the model's preference for the stereotypical sentence over the anti-stereotypical sentence.

To measure the overall bias of the model, the authors compute the percentage of pairs $(s_s,s_a)$ in the full CrowS-pairs dataset for which the model prefers the the stereotypical sentence $s_s$ over the anti-stereotypical $s_a$, i.e.,
\begin{equation}\label{eq: crowS-plog}
     B_{CrowS} := \frac{100}{N}\sum_{(s_s,s_a)}\mathbb{I}(pll(s_s)>pll(s_a))
\end{equation}
For the original English implementation, the number of pairs in the dataset is $N= 1508$.

\section{Conclusion and Outlook}
The BIAS Detection Framework provides the foundation to publish the open-source results from this Horizon Europe project to detect societal biases in word embeddings and language models. The BIAS project investigates several languages including German, Dutch, Icelandic, Norwegian, Turkish and Italian, and results will be added to the repository continuously. 

In the second half of the project, this work will be extended with the BIAS Mitigation Framework, providing insights into our investigations on reducing bias in word embeddings and language models for European languages.

\section*{Acknowledgments}
This work is part of the Europe Horizon project BIAS, grant agreement number 101070468, funded by the European Commission, and has received funding from the Swiss State Secretariat for Education, Research and Innovation (SERI).


\end{document}